\documentclass{article}

\usepackage[final]{nips_2017}

\usepackage[utf8]{inputenc} 
\usepackage[T1]{fontenc}    
\usepackage{hyperref}       
\usepackage{url}            
\usepackage{booktabs}       
\usepackage{amsfonts}       
\usepackage{nicefrac}       
\usepackage{microtype}      

\usepackage[pdftex]{graphicx}

\title{Embedded Real-Time Fall Detection Using Deep Learning For Elderly Care}

\author{
  Hyunwoo Lee\thanks{Alternative Email: lhw0772@gmail.com}
  \And
  Jooyoung Kim 
  \And
  Dojun Yang
  \And
  Joon-Ho Kim \\
  Samsung Research, Samsung Electronics \\
  \texttt{\{hyun0772.lee, joody.kim, dojun.yang, mythos.kim\}@samsung.com} \\
}

\begin{document}

\maketitle

\begin{abstract}
This paper proposes a real-time embedded fall detection system using a \textbf{DVS(Dynamic Vision Sensor)}(\citet{6884527}) that has never been used for traditional fall detection, a dataset for fall detection using that, and a \textbf{DVS-TN(DVS-Temporal Network)}.
The first contribution is building a \textbf{DVS Falls Dataset}, which made our network to recognize a much greater variety of falls than the existing datasets that existed before and solved privacy issues using the \textbf{DVS}.
Secondly, we introduce the \textbf{DVS-TN} : optimized deep learning network to detect falls using \textbf{DVS}.
Finally, we implemented a fall detection system which can run on low-computing H/W with real-time, and tested on \textbf{DVS Falls Dataset} that takes into account various falls situations.
Our approach achieved \textbf{95.5\%} on the F1-score and operates at \textbf{31.25 FPS} on NVIDIA Jetson TX1 board.
\end{abstract}

\section{Introduction}
As the elderly population grows, care for the elderly living alone is becoming an important issue. According to \citet{fuller2000falls}, falls are primary etiology of accidental deaths in persons over the age of 65 years in the United States and, especially in the case of an elderly person living alone, the risk increases even more. If there is no quick assistance after the fall then it can lead to a fatal result. Therefore it is important to monitor in their home and let emergency agency or family members to know that situation . There have been many studies on fall detection. Traditional fall detection systems can be divided into two major ways.

\textbf{Firstly, there are some studies using wearable sensors.}
\citet{chen2006wearable} used the magnitude of acceleration for fall
detection.\citet{pierleoni2015high} used an inertial unit including a triaxial accelerometer, 
gyroscope, and magnetometer, and applied effective data fusion. Most commercialized systems are adapting those methods and they have some advantages in the perspective of low cost and low power on embedded systems. However, there are limitations in accuracy, inconvenience of charging the sensor or wearing the sensor on the waist.

\textbf{Some studies used vision sensors to detect falls.}
\citet{yu2012posture} used a single camera to estimate the posture, features are extracted by an eclipse which is fitted to the human profile and projection histogram and then it is classified by SVM(Support Vector Machine). 
With vision sensors, you can see high accuracy in limited situations, but in exceptional situations such as occlusions and variation of light, recognition accuracy may not be good enough and there is also a possibility of privacy issue because it should be installed at home.
\citet{Feng2014DeepLF} applied a background extraction, blob-merging method, DBN (Depp Belief Network) and RBM (Restricted Boltzmann machine) to existing Vision Sensor Based System. In addition,\citet{wang2016temporal} proposed Temporal Segment Networks and it achieved 69.4\% in hmdb51 and 94.2\% in UCF101, respectively. But they focused on action recognition on video clips rather than real-time fall detection on live video stream. Even more, they need high-performance H/W to make it possible to detect fall in real-time.

So, in order to solve the above problems, we introduce a real-time fall detection system on embedded computing unit. We \textbf{(1)} built a dataset for fall detection using \textbf{DVS} to minimize a privacy problem in indoor environment, \textbf{(2)} introduce DVS-TN that can detect falls in real-time, and \textbf{(3)} implemented a fall detection system on NVIDIA Jetson TX1 board using \textbf{only CPU}.
\section{DVS-Dataset Construction}
\begin{table}[t]
  \caption{\textbf{DVS Falls Dataset.} falls were recorded at four directions (front,back,left,right).}
  \label{DVS Fall Dataset}
  \centering
  \small
  \begin{tabular}{lll}
    \toprule
    Action   & Description & Number of clips \\
    \midrule
    idle & stand and do nothing &62\\
    lie & lie down slowly &62\\
    walk & walk around &62\\
    sit & sit down &124\\
    stand & stand up after other actions &744\\
    fall & fall down & 558 \\
    fake fall& almost fall but stand up & 248 \\
    \bottomrule
  \end{tabular}
\end{table}
DVS data uses 1 channel and is similar to RGB Difference. Accumulated motion events for each pixel for a very short period of time preserve information about human shapes and movements. So it has the advantage of reducing privacy problem compared to RGB data.
There are no falls dataset suitable to train deep-learning model for us. Although there are few datasets, such as HMDB(\citet{Kuehne}), UCF-101(\citet{soomro2012ucf101}), and Activity-Net(\citet{caba2015activitynet}) that deal with overall action recognition, they contain some problems. One is that they are using images captured at various camera position. The viewpoint of \textbf{DVS} is required to be fixed so that it can focus on moving objects only. Another problem is that there is not enough data related to falls.

With the above problems in existing datasets, we have configured a dataset like Table~\ref{DVS Fall Dataset} to detect falls using DVS. DVS image is recorded at the 640x480 size and we use this image by resizing it for learning and testing our model. We captured 31 people walking, lying down, sitting on the floor, sitting on a chair, getting up and falling action. We collected 1860 DVS clips. Among them, we used 1500 clips for training and 360 clips for testing. In order to capture images from various angles, two cameras were installed at different angles and the falling action was taken in four directions such as front, back, left, and right. Our problem is the binary classification (fall down or non-fall) in real-time on embedded system. In our experiment, we found out that the average length of falls is about 50 frames. So, for the fall down data we extracted 50 frames from the recorded data. For the non-fall data, we gathered atomic actions as Table~\ref{DVS Fall Dataset} because every action can be considered as combination of atomic actions.
As a result, we built the DVS Fall Dataset in the form of atomic action which has 50 frames length for each action.
\section{Method}
\graphicspath{ {./} }
\begin{figure}[h]
  \centering
  \fbox{\rule[0cm]{0cm}{0cm} 
   \includegraphics[width=0.9\linewidth]{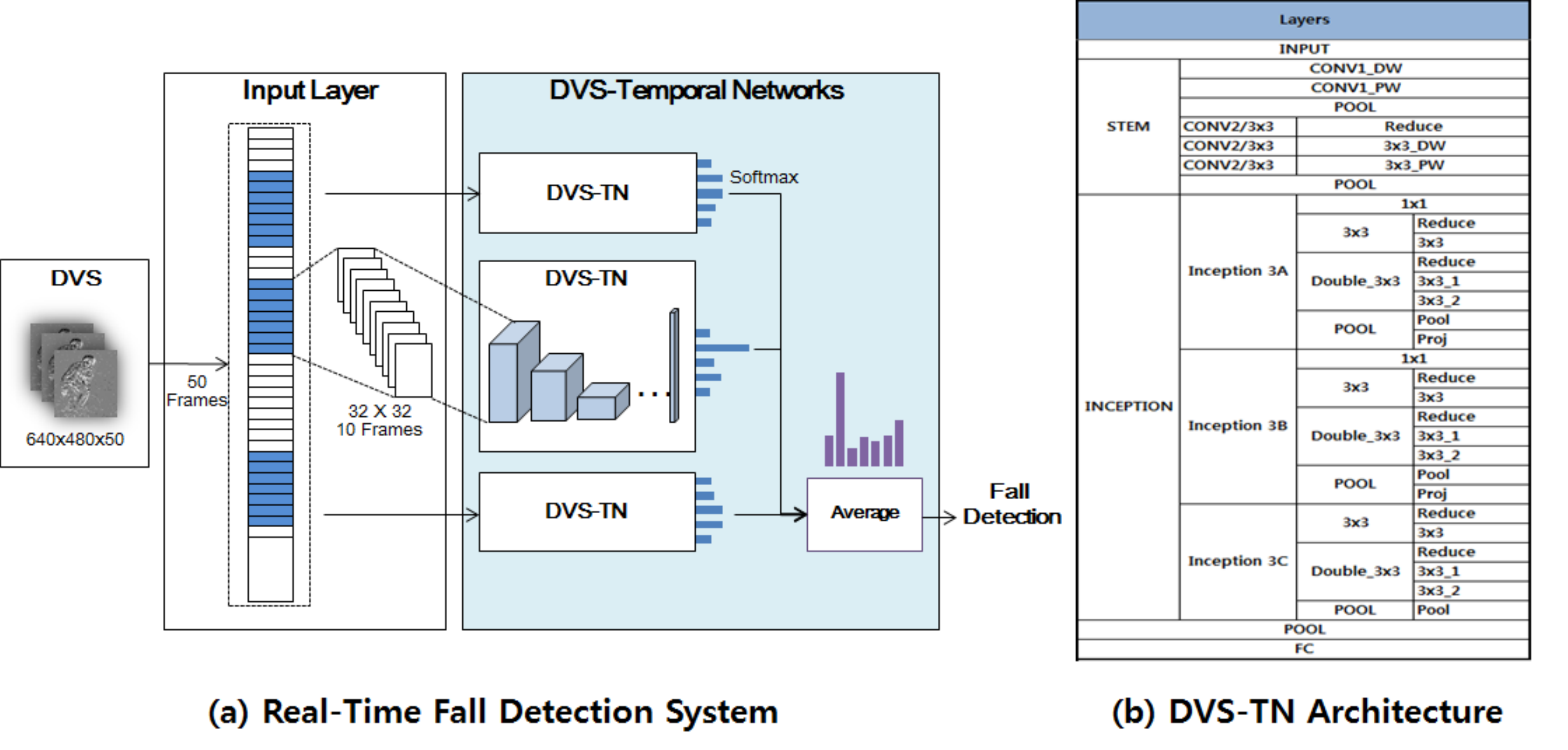}
  }
  \caption{Real-time Fall Detection System and Architecture of DVS-TN}
  \label{Architecture of DVS Temporal Network}
\end{figure}
We have modified \textbf{TSN} used by \citet{wang2016temporal}. \citet{wang2016temporal} firstly divides the video into K segments and a short snippet is randomly selected from each segment. And then each snippet is processed by Temporal-Convnet and Spatial ConvNet. Spatial ConvNet processes RGB images from snippets, Temporal ConvNet processes optical flow extracted from the RGB image. The class scores of different snippets are fused by an the segmental consensus function to yield segmental consensus, which is a video-level prediction. 
But when using the DVS image, the method of extracting the optical flow from DVS images is not appropriate because they already have shown the only different parts. And because \citet{wang2016temporal} showed reasonable performance using stacked RGB Difference as input, we used stacked DVS image similar to RGB difference as network input. And when using both Spatial ConvNet and Temporal ConvNet, experimental results show similar accuracy for our dataset compared with using only Temporal ConvNet. So we use only temporal network which uses stacked DVS images as shown at Figure~\ref{Architecture of DVS Temporal Network}. We choose bn-inception (inception v2) (\citet{ioffe2015batch}) for base model as \citet{wang2016temporal} did. 
\subsection{Model Optimization}
\graphicspath{ {./} }
\begin{figure}[h]
  \centering
  \fbox{\rule[0cm]{0cm}{0cm} 
   \includegraphics[width=0.9\linewidth]{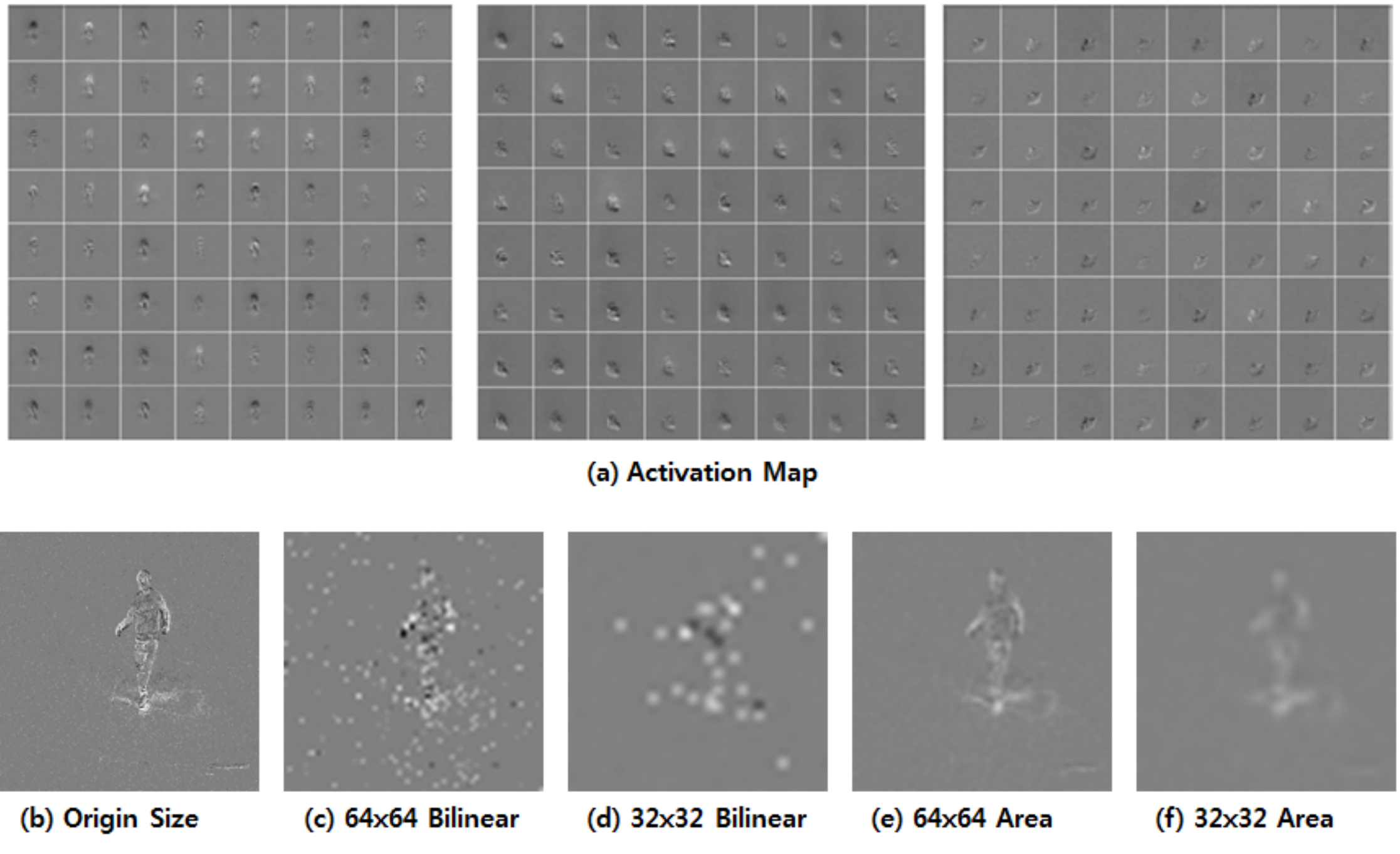}
  }
  \caption{\textbf{Activation Map(a)} The input image is based on the model with 128x128 size and it's an activation map for 10 frames per 1 segment in total 3 segments. \textbf{Comparision of resizing method (Bilinear Interpolation vs Area Interpolation) (b-f)} When the interpolation method is inter-area, it is possible to minimize the feature loss and to maintain the accuracy. }
  \label{act_and_resize}
\end{figure}
The most important thing for detecting falls is not high-quality features but overall shape of the body. Therefore, the input size and the output dimension of convolution can be minimized as long as the performance is maintained. We tested the recognition accuracy of the existing base network from 224x224 to 32x32 step by step. When the interpolation method is bi-linear, the recognition accuracy was maintained to 64x64 but decreased at 32x32. However by changing interpolation method to inter-area, it was possible to minimize the feature loss and to maintain the accuracy as shown at figure~\ref{act_and_resize}.
And figure~\ref{act_and_resize} shows the activation map for the first convolution layer of the stem using an extract feature tool from caffe. We can guess that a lot of similar filter patterns are learned. This can guarantee similar accuracy with channel reduction, which reduces the number of filters in each layer by 3/4.
In addition, we applied some optimization techniques to our network. We divide the existing Convolution Layer into pointwise convolution and depthwise convolution for model speed up as \citet{howard2017mobilenets} did. And the scale/shift parameter learned in the batch normalization layer can be processed at the convolution layer when run-time according to \citet{kim2016pvanet}. When applied to \textbf{DVS-TN}, there is a speed increase of 1.75 times compared with the previous one.
\subsection{Embedded Real-Time Fall Detection System}
We have enabled the \textbf{DVS-TN} to operate on the NVIDIA Jetson TX1 at \textbf{31.25 FPS} by applying the above-mentioned methods. We implemented a real-time fall detection system to run the optimized model on NVIDIA Jetson TX1. \footnote{Our real-time demonstration video is available at https://www.youtube.com/watch?v=RQkEDW2R7IA.}As shown in Figure~\ref{Architecture of DVS Temporal Network}(a) DVS-TN overview, our system is composed of two parts. The first resizes DVS images to 32x32 size, stacks them to 50 frames, and finally extracts each 10 frames which are front, middle and end segments of stacked frames. The second is \textbf{DVS-TN} that does classification about whether these three segments of 50 frames are fall or not.
\section{Experiment}
\begin{table}[t]
  \caption{\textbf{Model Compression Result.} 
  Accuracy is calculated based on F-1 score and speed is based on second using TX1 CPU. 
  M3 is modified model from bn-inception. 
  we reduced layer depth of bn-inception to speed up while maintaining accuracy. }
  \label{model compression}
  \centering
  \small
  \begin{tabular}{llllll}
    \toprule
    Model    & Input Dim & \# Parameters & \# Operations & F1-score & Speed \\
    \midrule
    M1 (TSN 224x224) & 224 & 10.4M & 2315.5M & 0.938 & 8.372\\
    M2 (64x64) & 64 & 10.4M & 189.0M & 0.954 & 0.767  \\
    M3 (v2stem+incep3) & 64 & 1.9M & 118.7M & 0.954 & 0.410\\
    M4 (M3+ seper conv) & 64 & 1.9M & 76.6M & 0.968 & 0.273 \\
    M5 (M4+ output depth reduct) & 64 & 1.1M & 45.5M & 0.963 & 0.175 \\
    M6 (M3+ 32x32) & 32 & 1.9M & 29.7M & 0.954 & 0.118\\
    M7 (32x32+ seper conv+ depth reduct) & 32 & 1.1M & 11.4M&0.955 &0.056 \\
    M8 (M7+ BN merge)& 32 &1.1M & 11.4M&0.955 & 0.047 \\
    \bottomrule
  \end{tabular}
\end{table}
\begin{table}[t]
  \caption{Model comparision between SOTA and Ours}
  \label{Model comparision between SOTA and Ours}
  \centering
  \tiny
  \begin{tabular}{lllllll}
    \toprule
    Model   &Input Type& Input Dimension & Speed(cpu) & Speed(gpu) & F1-score\\
    \midrule
    TSN Temporal-Net & Optical Flow& 224x224x10 & 8.372 & 0.275 & 0.938  \\
    Squeeze Net & DVS Image& 64x64x10 & 0.309 & 0.172 & 0.949  \\
    Ours & DVS Image & 32x32x10 & 0.047 & 0.032 & 0.955 \\
    \bottomrule
  \end{tabular}
\end{table}
\textbf{Model 5 and 6} show the results for input image size reduction and output dimension reduction. 
\textbf{Model 4} is experimental result for pointwise and depthwise Convolution. The number of operations decreased by 1.54 times when applying pointwise and depthwise convolution and it was actually 1.5 times faster on CPU. 
\textbf{Model 8} shows the effect of the technique of merging the scale and / shift learned by batch normalization at the convolution layer. This technique does not actually reduce the number of parameters in the network, but it reduces the run-time operation by merging the existing BN layer into the convolution layer.
Comparing the \textbf{model 8} with TSN as SOTA, the accuracy is nearly same based on F1-score and the execution speed increased by \textbf{178.12 times in CPU and 8.59 times in GPU.} 
\section{Conclusion}
We adapted the vision-based deep learning method, which is the most accurate method of the existing fall detection system. And we used the \textbf{DVS} to solve the existing privacy problem, constructed the falls dataset using it, and applied optimized deep learning and made the real-time fall detection system. We proved that a high accuracy fall detection system can be operated by \textbf{only CPU} on the embedded system with \textbf{DVS}. 
\bibliographystyle{plainnat}
\bibliography{mybib}
\end{document}